\documentclass{article} 

\usepackage{preprint}

\usepackage{amsmath, amsthm, amssymb, amsfonts}

\usepackage[numbers,square]{natbib}
\usepackage[utf8]{inputenc}	
\usepackage[T1]{fontenc}	
\usepackage{xcolor}		
\usepackage[colorlinks = true,
            linkcolor = purple,
            urlcolor  = blue,
            citecolor = cyan,
            anchorcolor = black]{hyperref}	
\usepackage{booktabs} 		
\usepackage{nicefrac}		
\usepackage{microtype}		
\usepackage{lineno}		
\usepackage{float}			

\usepackage{lipsum}		

\usepackage{newfloat}
\DeclareFloatingEnvironment[name={Supplementary Figure}]{suppfigure}
\usepackage{sidecap}
\sidecaptionvpos{figure}{c}

\usepackage{titlesec}
\titlespacing\section{0pt}{12pt plus 3pt minus 3pt}{1pt plus 1pt minus 1pt}
\titlespacing\subsection{0pt}{10pt plus 3pt minus 3pt}{1pt plus 1pt minus 1pt}
\titlespacing\subsubsection{0pt}{8pt plus 3pt minus 3pt}{1pt plus 1pt minus 1pt}

\usepackage{tikz,xcolor,hyperref}

\definecolor{lime}{HTML}{A6CE39}
\DeclareRobustCommand{\orcidicon}{
	\begin{tikzpicture}
	\draw[lime, fill=lime] (0,0) 
	circle [radius=0.16] 
	node[white] {{\fontfamily{qag}\selectfont \tiny ID}};
	\draw[white, fill=white] (-0.0625,0.095) 
	circle [radius=0.007];
	\end{tikzpicture}
	\hspace{-2mm}
}
\foreach \x in {A, ..., Z}{\expandafter\xdef\csname orcid\x\endcsname{\noexpand\href{https://orcid.org/\csname orcidauthor\x\endcsname}
			{\noexpand\orcidicon}}
}

\usepackage{adjustbox}
\usepackage{multirow}
\usepackage{draftwatermark}
\usepackage{afterpage}

\newcommand{\papername}{MV-MR}
\newcommand{\name}{MV--MR}

\title{\papername: multi-views and multi-representations for self-supervised learning and knowledge distillation}

\SetWatermarkText{\href{https://doi.org/10.3390/e26060466}{\color{gray}{Publication doi 10.3390/e26060466}}}
\SetWatermarkColor[gray]{0.6}
\SetWatermarkAngle{0}
\SetWatermarkScale{0.1}
\SetWatermarkHorCenter{300pt}
\SetWatermarkVerCenter{770pt}

\usepackage{authblk}

\author{Vitaliy Kinakh\orcidA}
\author{Mariia Drozdova\orcidB}
\author{Slava Voloshynovskiy\orcidC}

\affil{Department of Computer Science, University of Geneva}
\affil{\{ \textit{vitaliy.kinakh, mariia.drozdova, svolos}\} \textit{@unige.ch}}

\begin{document}

  \begin{@twocolumnfalse} 
  
\maketitle

\begin{abstract}
We present a new method of self-supervised learning and knowledge distillation based on multi-views and multi-representations (\papername). {\papername} is based on the maximization of dependence between learnable embeddings from augmented and non-augmented views, jointly with the maximization of dependence between learnable embeddings from the augmented view and multiple non-learnable representations from the non-augmented view. We show that the proposed method can be used for efficient self-supervised classification and model-agnostic knowledge distillation. Unlike other self-supervised techniques, our approach does not use any contrastive learning, clustering, or stop gradients. {\papername} is a generic framework allowing the incorporation of constraints on the learnable embeddings via the usage of image multi-representations as regularizers. The proposed method is used for knowledge distillation. {\papername} provides state-of-the-art self-supervised performance on the STL10 and CIFAR20 datasets in a linear evaluation setup. We show that a low-complexity ResNet50 model pretrained using proposed knowledge distillation based on the CLIP ViT model achieves state-of-the-art performance on STL10 and CIFAR100 datasets.
The code is available at: \href{https://github.com/vkinakh/mv-mr}{github.com/vkinakh/mv-mr}
\end{abstract}
\vspace{0.35cm}

  \end{@twocolumnfalse} 


\section{Introduction}
\label{sec:intro}
Self-supervised learning (SSL) methods are alternatives to supervised ones. In~recent years, the~gap between SSL and supervised methods has decreased in performing downstream tasks, including image classification~\cite{zhou2021ibot},  object detection~\cite{huang2022survey}, and semantic image segmentation~\cite{zheng2021hierarchical,punn2022bt}. A~general idea behind the SSL models for image classification is to train an embedding network, often called {an} {\em encoder},  on~an unlabeled dataset and then to use this pretrained encoder for the downstream~tasks.

The general goal is to ensure the invariance of embeddings to different inputs known as {\em{augmentations}} 
 or {\em {views}.} 
 However, this approach might lead to trivial solutions when two branches of encoders produce the same output. As~a result, one observes an effect known as a {\em collapse} in training when no meaningful representation can be learned for different inputs. Therefore, there have been a lot of recent works tackling this issue by regularizing the networks to avoid such a  collapse. Several key approaches have been developed to mitigate these negative impacts, using different tactics. The~first group of methods aims to directly maximize the mutual information between the input image $\bf x$ and its positive pairs created on the basis of augmentations. In~this view, an~exponential prior on the conditional distribution in the representation space and an associated contrastive loss with positive--negative pairs as in InfoNCE~\cite{oord2018representation} is assumed. Unfortunately, such an approach is quite computationally expensive in practice, due to the need for a large batch size to incorporate the large number of negative pairs. The~second group of methods aims to avoid collapse by introducing different asymmetries in two branches at the training stage. Examples of this approach are training one network with gradient descent and updating the other with an exponential moving average of the weights of the first network~\cite{chen2021exploring} or introducing regularizers on the learned representations such as regularization by decorrelation on the dimensions of the embeddings~\cite{zbontar2021barlow}, etc. The~third group of methods is Masked Image Modeling (MIM). These methods primarily focus on avoiding collapse and learning rich image representations by predicting the missing parts in masked inputs. This methodology relies on masking a portion of the input image and training the model to predict these masked parts, thereby learning contextual and semantic information. A~notable method in this domain is the BEiT~\cite{bao2021beit}, which introduces a transformer-based model that learns to predict the masked visual tokens, analogous to the masked language modeling in NLP. Another significant approach is the MAE (Masked Autoencoder) \cite{he2022masked}, which uses an asymmetric encoder--decoder structure, where the encoder processes only visible patches and the decoder reconstructs the masked patches. While MIM effectively learns representations, it is a transformer-specific approach and not transferable to other~architectures.

The proposed approach avoids the embeddings' collapse by introducing the dependence maximization between trainable embeddings and hand-crafted features using {\it {distance} 
 correlation} \cite{szekely2007measuring}. Distance correlation, unlike other losses in latent space, allows computing dependencies between feature vectors of different shapes. We maximize the dependence between different embeddings while preserving the variance in them. We show that variance preservation maximizes the entropy of embeddings, which makes them unique and distinguishable. Our approach is different from InfoNCE~\cite{oord2018representation}, which advocates a contrastive loss that maximizes the mutual information (MI) between input image $\bf x$ and its positive pairs. In~contrast to the InfoNCE, our approach is not contrastive, does not require large batch sizes, and~allows computing the distance between embeddings and features of any shape. It is also different from methods such as Barlow Twins~\cite{zbontar2021barlow} and VICReg~\cite{bardes2021vicreg} since we do not explicitly minimize the dependencies between the components within the~embedding.

We also show that the proposed approach can be used for efficient representation learning and latent space-agnostic knowledge distillation. The~approach is based on the dependence maximization between the embeddings of the target trainable encoder, represented by the ResNet50~\cite{he2016deep}, and~the embeddings of the pretrained encoder, represented by the CLIP~\cite{radford2021learning} (based on ViT-B-16~\cite{dosovitskiy2020vit}). Since the distance correlation is agnostic to the latent space shape, any pretrained encoder with any latent space can be used for knowledge distillation. To~our best knowledge, we are the first to propose a model-distillation method that is agnostic to the latent space~shape.

The main goal behind {\name} is twofold: (i) maximizing the invariance of embeddings, i.e.,~maximizing the proximity of embeddings for the same image observed under different views, and~(ii) maximizing the amount of information in each embedding, i.e.,~maximizing the variability of the embedding. Furthermore, to~avoid the collapse during training, we regularize the branch with the augmentations by imposing the dependence constraints on a set of representations extracted from various~encodings.

The proposed approach introduces several unique features: (i) we introduce a novel SSL approach that avoids collapse thanks to an additional regularization term that maximizes the dependence between trainable embeddings and various feature vectors using distance correlation; (ii) up to our best knowledge, the~proposed method is among the first that uses the dependence maximization of the latent space based on distance correlation for SSL; (iii) the proposed method is agnostic to the latent space shape and, thus, can be used with any types of features; (iv) we introduce a novel knowledge distillation technique that is agnostic to model and shape of latent space; (v) we demonstrate the state-of-the-art classification results on the STL10~\cite{coates2011analysis} (89.71\%) and CIFAR20~\cite{van2020scan} (73.2\%) datasets using a linear evaluation protocol for non-contrastive SSL methods; (vi) we provide the information-theoretic explanation of the proposed method that contributes to the explainable ML; (vii)~we demonstrate how the complex CLIP model with 86.2 M parameters trained on 400 M text--image pairs can be distilled into a ResNet50 model with just 23.5 M parameters trained on the STL10, CIFAR100~\cite{krizhevsky2009learning}, and ImageNet-1k~\cite{deng2009imagenet} datasets; (viii) we achieve state-of-the-art performance in knowledge distillation in the image-classification task using the ResNet50 model as student and CLIP ViT-B-16 as teacher on CIFAR100, with {\bf {78.6\%} } accuracy. 

We have three loss terms in our objective function: 
(a) the first term $\mathcal{L}_1$ consists of the mean square error (MSE) loss between the embeddings from the non-augmented view and augmented views of the same image; it is used for the invariance of embeddings, and~we introduce additional variation terms that are used for maximization of the variability of the embeddings (we demonstrate that this term originates from an upper bound on mutual information between these embeddings under corresponding assumptions); (b) the second term $\mathcal{L}_{2}$ stands for the distance correlation between the embeddings from the augmented and non-augmented views that complements the first term to capture non-linear relations between the embeddings; and (c) the third term $\mathcal{L}_{3}$ corresponds to the distance correlation between the embeddings from the augmented view and multiple image representations. For~the non-learnable or hand-crafted representations, we have studied various techniques of invariant data representation that are well-known in computer vision and image-processing applications. The~studied hand-crafted features include, but are not limited tp, ScatNet~\cite{oyallon2018scattering} features, local standard deviation (LSD)-based~\cite{narendra1981real} filters, and~histograms of oriented gradients (HOG) \cite{dalal2005histograms}. Additionally, to~demonstrate the flexibility of the proposed method, we have also considered random augmentations of the original images flattened into feature vectors as instances of hand-crafted features. Since distance correlation is shape-agnostic for the features, we are able to combine features of different shapes in the loss functions. Also, replacing hand-crafted features with embeddings from pretrained networks is used for model distillation, without~the need to change losses, architecture, or feature~dimensionality.

\section{\papername: Motivation and~Intuition}
\begin{figure} 
   \centering
    \includegraphics[width=0.9\textwidth]{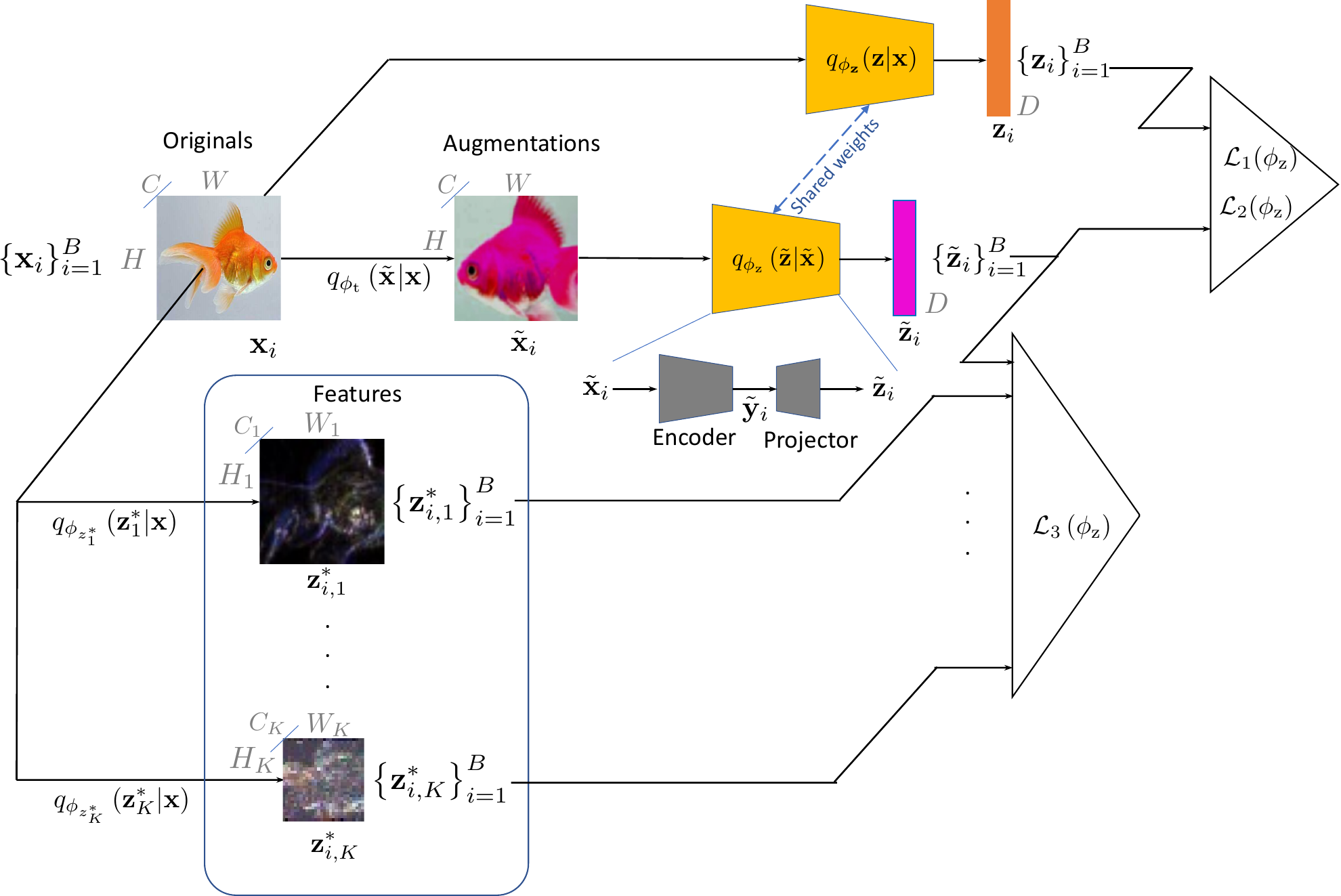}
    \caption{
    \textls[-15]{\papername: proposed SSL approach.  Two {\em views} of the image are produced: one original and the other augmented by $q_{\phi_{\mathrm{t}}}(\tilde{\mathbf{x}}| \mathbf{x})$. Then, this first view is encoded via encoder $q_{\phi_{\mathbf{z}}}(\mathbf{z}|\mathbf{x})$, producing ${\bf z}_i$, which denotes an original embedding, and via $q_{\phi_{\mathbf{z}}}(\tilde{\mathbf{z}}|\tilde{\mathbf{x}})$, producing $\tilde{\mathbf{z}}_{i}$, denoting an augmented one. The~representations $\mathbf{z}_{i, k}^{*}$ are obtained via $K$ hand-crafted feature extraction mappers $q_{\phi_{z_{k}^{*}}}\left(\mathbf{z}_{k}^{*} | \mathbf{x}\right), 1\leq k \leq K$. The~same process is applied to each image ${\bf x}_i$ in the batch $1 \leq i \leq B$.
    The embedding is regularized by a loss $\mathcal{L}_1(\phi_{\mathrm{z}})$, minimizing the Euclidean distances between the embeddings ${\bf z}_i$ and $\tilde{\mathbf{z}}_{i}$ while ensuring that their variance is above a threshold. The~loss $\mathcal{L}_2(\phi_{\mathrm{z}})$ ensures the dependence between the pair of augmented and non-augmented embeddings using the distance correlation. The~regularization loss $\mathcal{L}_3(\phi_{\mathrm{z}})$ is imposed by maximizing the distance correlation between the augmented embedding $\tilde{\mathbf{z}}_{i}$ and a set of hand-crafted features $\mathbf{z}_{i, k}^{*}, 1 \leq k \leq K$ computed for the given batch $B$.}}
    \label{fig:d_cor_ssl}
\end{figure}

\begin{figure} 
   \centering
    \includegraphics[width=0.9\textwidth]{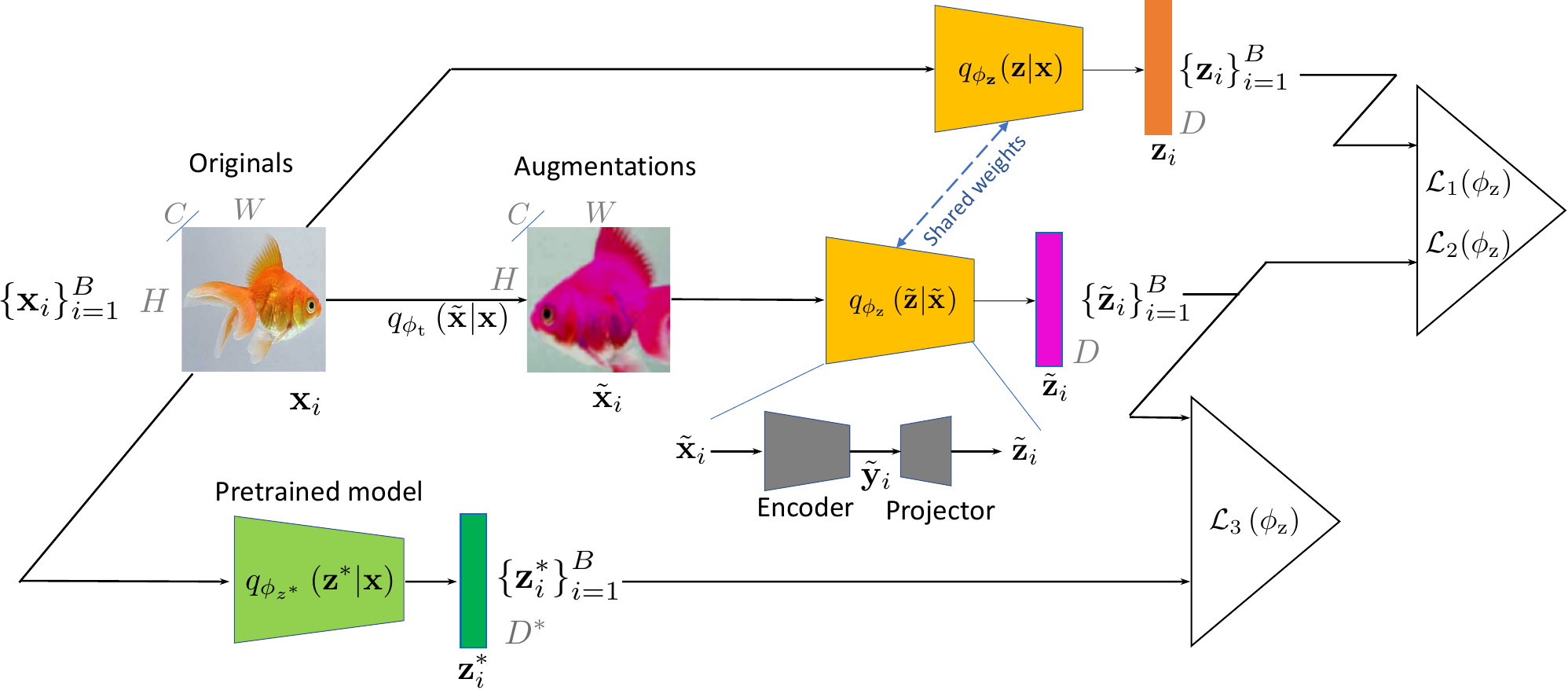}
    \caption{\textls[-15]{\papername: distillation approach. $q_{\phi_{z^{*}}}(\textbf{z}^{*}|\textbf{x})$ is the high-complexity (in term of parameters) teacher model used as a feature extractor in order to train a low-complexity student model $q_{\phi_{z}}\left(\mathbf{z} | \mathbf{x}\right)$. The~teacher model corresponds to a set of  hand-crafted feature extractors in Figure~\ref{fig:d_cor_ssl}. The~representations $\mathbf{z}_{i}^{*}$ are obtained from the pretrained teacher model $q_{\phi_{z^{*}}}(\textbf{z}^{*}|\textbf{x})$. The~same losses as in self-supervised pretraining are used: $\mathcal{L}_1(\phi_{\mathrm{z}})$ minimizes the Euclidean distances between the embeddings ${\bf z}_i$ and $\tilde{\mathbf{z}}_{i}$ while ensuring that their variance is above a threshold, $\mathcal{L}_2(\phi_{\mathrm{z}})$ ensures the dependence between the pair of augmented and non-augmented embeddings using the distance correlation, and $\mathcal{L}_3(\phi_{\mathrm{z}})$ maximizes the distance correlation between the augmented embedding $\tilde{\mathbf{z}}_{i}$ and the teacher's embeddings $\mathbf{z}_{i}^{*}$.}}
    \label{fig:mv_mr_distill}
\end{figure}

\textls[-22]{{\papername} pretraining and distillation schemes are schematically shown in \mbox{Figures~\ref{fig:d_cor_ssl} and \ref{fig:mv_mr_distill}}}, respectively. The~dimensions of embeddings with and without augmentations are the same, i.e.,~$\tilde{\bf{z}} \in \mathbb{R}^D$ and ${\bf z} \in \mathbb{R}^D$, respectively. These embeddings are extracted from the augmented $\tilde{\bf x}$ and non-augmented $\bf{x}$ via a generalized parametrized embedder  $q_{\phi_{z}}(\cdot | \cdot)$ that can be deterministic or stochastic with parameters $\phi_{z}$. The~encoder can be a parametrized neural network of any architecture. A~$k^{th}$  hand-crafted descriptor ${\bf{z}}_k^*$, where~$k\in\{1,2,\cdots, K \}$ and $K$ stands for the total number of hand-crafted descriptors, is generally a tensor of dimensions $H_k \times W_k \times C_k$ and is flattened to $D_k = H_k  W_k C_k$. This descriptor is generally obtained via deterministic assignment  ${\mathbf{z}_{k}^{*}} = f_{\phi_{z_{k}^{*}}}({\bf x})$ or sometimes via stochastic mapping $ {\bf Z}^*_k \sim q_{\phi_{z_{k}^{*}}}\left(\mathbf{z}_{k}^{*}| \mathbf{x}\right)$, where $\phi_{{\bf{z}^{*}}_k}$ denotes the parameters of the $k^{th}$ feature~extractor.

\subsection{Motivation: Regularization in Self-Supervised Representation~Learning}
The learned representation should contain the informative representation of data with lower dimensionality and should be invariant under some transformations, i.e.,~to ensure the same latent representation for the data from the same sample passed through certain transformations. The~satisfaction of these conflicting requirements in practice is not a trivial task. Many state-of-the-art SSL techniques try to find a reasonable compromise between these requirements and practical feasibility solely in the scope of machine learning formulation by imposing certain constraints on the properties of the learned representation via the optimization of encoder parameters under~{augmentations.} 

At the same time, there exists a rich body of achievements in the computer vision community in the domain of the hand-crafted design of robust, invariant, yet discriminating data representations~\cite{loew2004distinctive,rublee2011orb,pietikainen2015two,dalal2005histograms}. Generally, the~computer vision descriptors are very rich in terms of targeted invariant features and quite efficient in terms of computation. However, to~our best knowledge, such descriptors are not yet fully integrated into the development of SSL techniques. Therefore, one of the objectives of this paper is to propose a framework where the SSL representation learning might be coupled with the constraints on the embedding space offered by the invariant computer vision representations. Our objective is not to consider a case-by-case approach on how to couple SSL with a particular computer vision representation but instead to propose a {\it{generic} 
 approach} where any form of desirable computer vision representation can be integrated into the SSL optimization problem in an easy and tractable way. This ensures that the learned representation possesses the targeted properties inherited from the used computer vision descriptors. Furthermore, features extracted by such descriptors might be considered as a form of invariant data representation, which is one of the desired properties of trained encoders. Thus, maximizing the dependence between the trainable embedding and such representation might be a novel form of regularization, leading to an increased-invariance yet collapse-avoiding technique. Since a single computer vision descriptor might not capture all desirable properties and have different representation formats, the~targeted framework should be flexible enough to deal uniformly with all these descriptors within a simple optimization problem. Distance correlation is very useful for this kind of representation learning, since it allows one to incorporate features of any shapes, without~the need to match the shape of learnable embeddings and hand-crafted target~embeddings.

In summary, our motivation is to include regularization constraints on the solution by borrowing some inherent computer vision feature invariance to certain transformations. In~this way, we target learning the low-dimensional embedding, which contains only essential information about the data that might be of interest for the targeted downstream task and where all information about the augmentations is~excluded.
\raggedbottom

\subsection{Intuition}
The basic idea behind {\papername} is to introduce constraints on the invariance of embedding via a new loss function. Our overall objective is to maximize the mutual information  $I(\tilde{\bf Z}; {\bf Z} )$ between the augmented embedding $\tilde{\bf Z}$ and the embedding without the augmentation  $\bf Z$ and to maximize the mutual information $I(\tilde{\bf Z}; {\bf Z}_k^* )$ between $\tilde{\bf Z}$  and some invariant feature ${\bf Z}_k^* $ extracted from $\bf X$ using a mapper that ensures a known invariance to the desired~transformation. 

\subsubsection{Measuring Dependencies between Embeddings of Non-Augmented and Augmented~Data}
\label{upper_bound_MI}

{\bf {Upper bound}  on mutual information}: In the first case, one can decompose the mutual information as
\begin{equation}
    \begin{aligned}
     I(\tilde{\bf Z} ; {\bf Z}) &= \mathbb{E}_{p(\tilde{\bf z},{\bf z})} \left[{\log \frac{p({\tilde{\bf z}}, {\bf z})}{p(\tilde{\bf z})p({\bf z})}}\right] &= \mathbb{E}_{p(\tilde{\bf z},{\bf z})} \left[{\log \frac{p({\tilde{\bf z}}| {\bf z})}{p(\tilde{\bf z})}}\right] = h(\tilde{\bf Z}) - h(\tilde{\bf Z}|{\bf Z}),
    \label{equation:MI_defintion}
     \end{aligned}
\end{equation}
where $h(\tilde{\bf Z}) = - \mathbb{E}_{p(\tilde{\bf z})} \left [ \log {p(\tilde{\bf z})}\right]$ denotes the differential entropy and $h(\tilde{\bf Z}|{\bf Z}) =  -  \mathbb{E}_{p(\tilde{\bf z},{\bf z})} \left[{ \log p({\tilde{\bf z}}| {\bf z})}\right]$ denotes conditional differential entropy
(we assume that the differential entropy is non-negative under the considered settings).
Since the computation of the marginal distribution $p(\tilde{\bf z})$ and conditional distribution $p({\tilde{\bf z}}| {\bf z})$  is difficult in practice, we proceed by bounding these terms. We assume that the desired embeddings need to be bounded by some variance $\sigma_z^2$ to avoid a training collapse when the encoders produce constant and non-informative vectors so that the entropy-maximizing distribution for the first entropy term is the Gaussian one, i.e.,~ $p(\tilde{\bf z}) \propto \frac{\exp \left(-\frac{1}{2}\mathbf{\tilde{z}}^{\mathrm{T}} \mathbf{\Sigma}_{\mathbf{z}}^{-1} \mathbf{\tilde{z}} \right)}{\sqrt{(2 \pi)^D|\mathbf{\Sigma}_{\mathbf{z}}|}}$, where $\mathbf{\Sigma}_{\mathbf{z}}$ represents the covariance~matrix.

The conditional entropy $h(\tilde{\bf Z}|{\bf Z})$ is minimized when the embedding $\tilde{\bf Z}$ contains as much information as possible about ${\bf Z}$, i.e.,~when two vectors are dependent. Assuming that $p(\tilde{{\bf z}}| {\bf z}) \propto \frac{1}{C_z} \text{exp}(-\beta_{z} d(\tilde{{\bf z}}, {\bf z}))$, where $d(\tilde{{\bf z}}, {\bf z})$ denotes some distance between two vectors such as the $\ell_2$-norm for the Gaussian distribution or $\ell_1$-norm for the Laplacian one, where $C_z$ stands for the normalization constant and $\beta_z$ denotes a scaling parameter. Thus, the~minimization of the conditional entropy $h(\tilde{\bf Z}|{\bf Z})$ reduces to the minimization of the distance $d(\tilde{{\bf z}}, {\bf z})$.

{\bf {Distance} 
 covariance}: Another way to measure the dependency between the data is based on distance covariance, as proposed by~\cite{szekely2007measuring}. In~the general case of dependence between the data, the~distance covariance is non-invariant to strictly monotonic transformations, unlike mutual information. Nevertheless, the~distance covariance has several attractive properties: (i) it can be efficiently computed for two vectors that have generally different dimensions $\tilde{\bf z} \in \mathbb{R}^D$ and ${\bf z} \in \mathbb{R}^{D'}$, such that $D \neq D'$, and (ii) it is easier to compute in practice in contrast to the mutual information. Additionally, the~distance covariance captures higher-order dependencies between the data, in contrast to the Pearson correlation.
The {\em distance covariance} $\operatorname{dCov}^2(\tilde{\mathbf{Z}}, \mathbf{Z})$, proposed by~\cite{szekely2007measuring}, is defined as
\begin{equation}
    \label{equation:d_cov_analytical}
    \operatorname{dCov}^2(\mathbf{Z}, \tilde{\mathbf{Z}})=\frac{1}{c_D c_{D'}} \int_{\mathbb{R}^{D+D'}} \frac{\left|\varphi_{\mathbf{Z}, \tilde{\mathbf{Z}}}(\mathbf{t}, \mathbf{u})-\varphi_{\mathbf{Z}}(\mathbf{t}) \varphi_{\tilde{\mathbf{Z}}}(\mathbf{u})\right|^2}{|\mathbf{t}|_D^{1+D}|\mathbf{u}|_{D'}^{1+D'}} d \mathbf{t} d \mathbf{u},
\end{equation}
which measures the distance between the joint characteristic function  $\varphi_{\mathbf{Z}, \tilde{\mathbf{Z}}}(\mathbf{t}, \mathbf{u})$ and the product of the marginal characteristic functions $\varphi_{\mathbf{Z}}(\mathbf{t}) \varphi_{\tilde{\mathbf{Z}}}(\mathbf{u})$ \cite{szekely2007measuring}. This definition has a lot of similarities to the mutual information in (\ref{equation:MI_defintion}), which measures the ratio between the joint distribution $p({\tilde{\bf z}}, {\bf z})$ and the product of marginals $p(\tilde{\bf z})p({\bf z})$.
Since $\varphi_{\mathbf{Z}, \tilde{\mathbf{Z}}}(\mathbf{t}, \mathbf{u}) = \varphi_{\mathbf{Z}}(\mathbf{t}) \varphi_{\tilde{\mathbf{Z}}}(\mathbf{u})$ when $\tilde{\mathbf{Z}}$ and $\mathbf{Z}$ are independent random vectors, the distance covariance is equal to~zero.

In the following, we proceed with the normalized version of distance covariance, known as {\em distance correlation}, defined as
\begin{equation}
    \label{equation:d_corr}
 \operatorname{dCor}(\tilde{\textbf{Z}}, \textbf{Z})=\frac{\mathrm{dCov}^{2}(\tilde{\textbf{Z}},\textbf{Z})}{\sqrt{\mathrm{dVar}(\tilde{\textbf{Z}}) \mathrm{dVar}(\textbf{Z})}},
\end{equation}
where $0 \leq \operatorname{dCor}(\tilde{\textbf{Z}}, \textbf{Z}) \leq 1$ and $\operatorname{dVar}(\textbf{Z}) = \operatorname{dCov}^{2}(\textbf{Z}, \textbf{Z})$.

{\em Sample distance covariance}, for a given
$\textbf{Z}_B = [\textbf{z}_{1}, ..., \textbf{z}_{B}]$, denoting a batch of size $B$ of embeddings from original views, and~ $\tilde{\textbf{Z}}_B = [\tilde{\textbf{z}}_{1}, ..., \tilde{\textbf{z}}_{B}]$, referring to a batch of embeddings from augmented views, is defined as
\begin{equation}
    \label{equation:dcov}
    \mathrm{dCov}_{B}^{2}(\tilde{\textbf{Z}}_B, \textbf{Z}_B):=\frac{1}{B^{2}} \sum_{j=1}^{B}\sum_{i=1}^{B} A_{j, i} C_{j, i}.
\end{equation}
In Equation~(\ref{equation:dcov}), we use the notations $A_{j, i}:=a_{j, i}-\bar{a}_{j \cdot}-\bar{a}_{\cdot i}+\bar{a}_{\cdot \cdot}, \quad$ $C_{j, i}:=c_{j,i}-\bar{c}_{j \cdot}-\bar{c}_{\cdot i}+\bar{c}_{\cdot \cdot}$,  where $\quad a_{j, i} =\left\|\tilde{\textbf{Z}}_{B_j}-\tilde{\textbf{Z}}_{B_i}\right\| , \quad$ $c_{j, i} =\left\|\textbf{Z}_{B_j}-\textbf{Z}_{B_i}\right\|$, where $j, i=1,2, \ldots, B$. Finally, {\em sample distance correlation} is defined as:
\begin{equation}
    \label{equation:sample_d_corr}
 \operatorname{dCor_B}(\tilde{\textbf{Z}}_B, \textbf{Z}_B)=\frac{\mathrm{dCov_B}^{2}(\tilde{\textbf{Z}}_B,\textbf{Z}_B)}{\sqrt{\mathrm{dVar}_{B}(\tilde{\textbf{Z}}_{B}) \mathrm{dVar}_{B}(\textbf{Z}_B)}},
\end{equation}
with $\mathrm{dVar}_{B}(\textbf{Z}_B) = \mathrm{dCov}_{B}^{2}({\textbf{Z}_B}, \textbf{Z}_B)$.

\subsubsection{Dependence between Embeddings of Augmented Data and Multiple Hand-Crafted~Representations}

The second mutual information $I(\tilde{\bf Z} ; {\bf Z}_k^* )$ between $\tilde{\bf Z}$  and some invariant feature ${\bf Z}_k^* $ deals with vectors of different dimensions. Thus, one can either map these vectors to the same dimension and apply the above arguments, use the Hilbert--Schmidt proxy~\cite{NIPS2007_d5cfead9}, or~proceed with the distance correlation dependence measure for the uniformity of consideration. We focus on the distance correlation case due to its property of handling vectors of different dimensions and its ability to capture higher-order data statistics. 

\section{Related~Work}
{\bf {Pretext} 
 task methods.} The main idea behind these methods is to design a specific task, {\em a.k.a. pretext task}, for the dataset that contains some ``labels'' of the pretext task without having any access to the labels of the target task. Such pretext tasks include, but are not limited to, applying and predicting parameters of the geometric transformations~\cite{gidaris2018unsupervised}, jigsaw puzzle solving~\cite{noroozi2016unsupervised}, inpainting~\cite{pathak2016context} and colorization~\cite{larsson2017colorization} of the images, and reversing augmentations. Typically, the~pretext task methods have been coupled with other SSL techniques in recent years~\cite{kinakh2021scatsimclr,yi2022using,zaiem2021pretext}.

{\bf {Contrastive} 
 methods.} Most of the contrastive SSL methods are based on different extensions of the InfoNCE~\cite{oord2018representation} formulation. 
The InfoNCE method is based on the direct maximization of the mutual information between the input image and its positive pairs via minimization of the contrastive loss. Examples of contrastive methods are SimCLR~\cite{chen2020simple}, SwAV~\cite{caron2020unsupervised}, and DINO~\cite{caron2021emerging}.

{\bf {Clustering} 
 methods.} Clustering-based SSL methods are based on the idea of assigning cluster labels to the learned representations in an unsupervised manner with some regularization, such as maintaining uniformity of these cluster labels. The DeepCluster~\cite{caron2018deep} method iteratively groups the features from the encoder using the standard {\it k}-means clustering and then uses them as an assignment for the supervision to update the weights of the encoder at the next iterations. SwAV~\cite{caron2020unsupervised} and DINO~\cite{caron2021emerging} are other notable clustering-based SSL methods that combine contrastive learning and clustering by clustering the data while requiring the same cluster assignment for different views of the same~image.

{\bf {Distillation} 
 methods.} Distillation-based SSL methods like BYOL~\cite{grill2020bootstrap}, SimSiam~\cite{chen2021exploring}, and others use the teacher--student type of training, where the student network is trained with the gradient descent, while the teacher network is not updated with gradient descent, but~rather with an exponential moving-average update or other method. Such a design is used to avoid~collapse. 

{\bf {Collapse}-
 preventing methods.} Similar to distillation, collapse-preventing methods try to prevent the collapse by the usage special regularization of embeddings. The Barlow Twins~\cite{zbontar2021barlow} method aims to make the covariance matrix of the embeddings to be an identity matrix. This means that each dimension of the embeddings should be decorrelated with all other dimensions. Additionally, the~minimum variance of embedding per each dimension in the batch is constrained. The VICReg~\cite{bardes2021vicreg} method extends the Barlow Twins~\cite{zbontar2021barlow} approach by imposing an additional constraint on the distance between the embeddings with and without~augmentations. 

 {\bf {Masked} 
 Image Modeling.} Masked Image Modeling (MIM) for self-supervised learning has emerged as a compelling approach, diverging from traditional methods like pretext task, contrastive, or~clustering methods. Central to MIM is the principle of intentionally masking portions of an input image and training a model to predict these occluded parts. This process enables the model to learn valuable representations of the data without relying on explicit labels. Unlike contrastive learning methods like SimCLR or SwAV that require negative samples, MIM directly utilizes the spatial coherence of images to enhance the model's ability to recognize and predict the structure within masked areas. Pioneering examples include the BEiT~\cite{bao2021beit} algorithm, which employs a transformer architecture to predict the masked visual tokens, drawing inspiration from masked language modeling. Another notable implementation is the MAE (Masked Autoencoder) \cite{he2022masked}, which uses an asymmetric encoder--decoder structure to efficiently reconstruct masked patches. These approaches contrast with distillation methods like BYOL, where a teacher--student model is used, and~clustering methods like DeepCluster that focus on feature clustering. MIM's uniqueness lies in its direct engagement with the raw image data, offering a pathway to learn intricate image features in a self-supervised manner without the need for complex negative sample handling or clustering~mechanisms.

{\bf {Knowledge}
 distillation.} Knowledge distillation~\cite{gou2021knowledge} is a type of model optimization, where a simple small model (student model) is trained to match the bigger complex model (teacher model). There are multiple types of knowledge-distillation schemes: offline distillation~\cite{hinton2015distilling},  online distillation~\cite{mirzadeh2020improved}, and~self-distillation~\cite{zhang2019your}. There are multiple types of knowledge types that are used for distillation: response-based knowledge~\cite{hinton2015distilling,ba2014deep}, feature-based knowledge~\cite{romero2014fitnets}, and~others. We show how our method can be used for offline feature-based knowledge distillation.
\section{\papername: Detailed~Description}
\unskip

\subsection{Method}
The training objective consists of two parts: (a) ensuring the invariance of the learned representation under the applied augmentations and, simultaneously, (b) imposing constraints on the learned representation. The~final loss integrates the loss based on the upper bound of the mutual information and distance~correlation.

\subsubsection{Training Objectives for the Representation Learning Based on the Mutual~Information}

We follow the overall idea of ensuring the invariance of learned representation under the family of applied augmentations and we proceed along the line discussed in the previous section. Since both branches have the same dimension $D$, we proceed with the maximization of the upper bound on the mutual information between these dimensions, as considered in Section~\ref{upper_bound_MI}. 

To train the deterministic encoder $q_{\phi_{z}}(\mathbf{z}| \mathbf{x}) = \delta(\mathbf{z} - f_{\phi}(\mathbf{x}))$, we penalize the embeddings $\tilde{\mathbf{z}}_{i}$ of augmented view $\tilde{\mathbf{x}}_{i}$ to be as close as possible to the embeddings $\mathbf{z}_{i}$ of non-augmented view $\mathbf{x}_{i}$ using the MSE loss between them:
\begin{equation}
    \label{equation:MSE}
    d\left(\mathbf{Z}_B, \tilde{\mathbf{Z}}_B\right)=\frac{1}{B} \sum_{i=1}^{B}\left\|\mathbf{z}_{i}-\tilde{\mathbf{z}}_{i}\right\|_{2}^{2}.
\end{equation}
The MSE is frequently used to ensure the similarity between embeddings. It can be demonstrated that this term equates to the conditional entropy term in the mutual information, as~specified in Equation~(\ref{equation:MI_defintion}), assuming Gaussian conditional distribution. At~the same time, it can be proven that InfoNCE aims to minimize the negative cross-entropy $h(\tilde{\bf Z}|{\bf Z})$ while maximizing the entropy $h({\bf Z})$ for the entropy-based model parametrization of $p(\tilde{\textbf{z}}|\textbf{z})$, with~$p(\tilde{\textbf{z}}) = \mathbb{E}_{p(\textbf{z})} \left[ p(\tilde{\textbf{z}} | \textbf{z}) \right]$. Hence, the~MSE loss is a non-contrastive loss based on $h(\tilde{\bf Z}|{\bf Z})$, while InfoNCE operates as its contrastive~counterpart.

The variance-regularization term corresponding to the entropy term in the mutual information in (\ref{equation:MI_defintion}) is used to control the variance of the embeddings. We use a hinge function of the standard deviation of the embeddings along the batch dimension:
\begin{equation}
    \label{equation:Variance_reg}
   v(\mathbf{Z}_B)=\frac{1}{D} \sum_{d=1}^{D} \max \left(0, \gamma-S\left(\textbf{z}[d], \epsilon\right)\right),
\end{equation}
\textls[-15]{where $\mathbf{z}[d]$ denotes the $d^{th}$ dimension of $\bf z$, $\gamma$ is the margin parameter for the standard deviation, $\epsilon$ is a small scalar for numerical stability, and $S$ is the standard deviation, defined as}
\begin{equation}
    \label{equation:STD}
    S(a, \epsilon)=\sqrt{\operatorname{Var}(a)+\epsilon}.
\end{equation}

We define the loss that ensures the correspondence between the embeddings from the augmented and non-augmented views, i.e.,~positive pairs. Simultaneously, we bound the variance of both embeddings as follows:
\begin{equation}
    \label{equation:loss_1}
    \mathcal{L}_{1}(\phi_{z}) = \lambda d\left(\textbf{Z}_B, \tilde{\textbf{Z}}_B \right)+\mu\left[v(\textbf{Z}_B)+v\left(\tilde{\textbf{Z}}_B\right)\right],
\end{equation}
where $\lambda$ and $ \mu$ are hyper-parameters controlling the importance of each term in the loss, $\mathbf{z} = f_{\phi_{\mathbf{z}}}(\mathbf{x})$ and $\tilde{\mathbf{z}} = f_{\phi_{\mathbf{z}}}(\tilde{\mathbf{x}})$. We set $\lambda$ and $\mu$ to 1 in our experiments. This loss is parametrized by the parameters $\phi_z$ of the encoder and projector. It should be pointed out that, for the symmetry, we impose the constraint on the variance for both augmented and non-augmented~embeddings.

It is interesting to point out that the obtained result coincides with the loss used in VICReg~\cite{bardes2021vicreg}, where its origin was not considered, from the information-theoretic point of view, as the maximization of mutual information between the embeddings. At~the same time, it should be noted that there is no constraint on the covariance matrix of embeddings as in the VICReg~\cite{bardes2021vicreg} and Barlow Twins~\cite{zbontar2021barlow} methods. 

\subsubsection{Training Objectives for Representation Learning Based on Distance~Covariance}

The distance correlation is used for the dependence maximization between the embedding from the augmented view with the non-augmented one and the set of representations from the hand-crafted~functions. 

Accordingly, the~loss
$\mathcal{L}_{2}(\phi_{z})$ denotes the minimization formulation of the distance correlation maximization problem between embeddings from augmented and non-augmented views:
\begin{equation}
    \label{equation:loss_2}
    \mathcal{L}_{2}(\phi_{z}) = \alpha \left[1 - \mathrm{dCor_B}(\tilde{\textbf{Z}}_B, {\textbf{Z}_B}) \right],
\end{equation}
and the loss $\mathcal{L}_{3}(\phi_{z})$ denotes the same for  the embedding from the augmented view and $k^{th}$ hand-crafted representation (in the case of self-supervised pretraining) or embeddings from pretrained models (in the case of knowledge distillation):
\begin{equation}
    \label{equation:loss_3}
    \mathcal{L}_{3}(\phi_{z}) = \sum_{k=1}^{K} {\beta_{k} \left [ 1 - \mathrm{dCor_B}(\tilde{\textbf{Z}}_B, \textbf{Z}_{B_k^{*}}) \right ]},
\end{equation}
where $\alpha$ and $\beta_{k}$ are hyper-parameters controlling the importance of each term in the loss. In~our experiments, we set $\alpha=1$ and $\beta_{k}=1$.

The final loss function is a sum of three losses:
\begin{equation}
    \label{equation:final_loss}
    \mathcal{L}\left( \phi_{z} \right)=\mathcal{L}_{1}(\phi_{z}) + \mathcal{L}_{2}(\phi_{z}) + \mathcal{L}_{3}(\phi_{z}).
\end{equation}

\section{Applications of the Proposed~Model}

\textls[-15]{In this section, we demonstrate the application of the proposed model to a) self-supervised pretraining of the model for classification and b) self-supervised model~distillation.}

\subsection{Self-Supervised Pretraining for~Classification}

The proposed method can be used for efficient self-supervised model pretraining for classification. Once pretrained, the~model is finetuned for classification. We report our results on the STL10~\cite{coates2011analysis} and ImageNet-1K~\cite{deng2009imagenet} datasets on linear evaluation and semi-supervised finetuning, with 1\% and 10\% of label pipelines in Tables~\ref{table:ImageNet_1k_evaluation} and~\ref{table:stl10_evaluation}. In~a linear evaluation pipeline, the~pretrained encoder is used as is, without~further training, while only a one-layer linear classifier is trained with labeled data. In~the semi-supervised finetuning pipeline, the~classifier head is attached to the pretrained encoder and the full model is finetuned on the labeled~data.

\subsection{Knowledge~Distillation}

The proposed method can also be used for efficient knowledge distillation. This is performed by using the pretrained model (teacher) as the feature extractor and computing the distance correlation between the embeddings of the trainable encoder (student) and the pretrained teacher encoder. In~practice, this can be used to match the performance of the big pretrained models with smaller models or match the performance of the models that have been trained on proprietary datasets. In~contrast to the standard knowledge distillation approaches~\cite{hinton2015distilling}, our approach does not use any labels or~require a latent space of the same shape. As~a practical example, we demonstrate that, by using the proposed knowledge distillation approach, we are able to match the performance of the CLIP~\cite{radford2021learning} based on ViT-B-16~\cite{dosovitskiy2020vit} with 86.2 M parameters pretrained on 400 M images from the LAION-400M~\cite{schuhmann2021laion} dataset, using ResNet50 with only {\bf {23.5 M} 
} parameters pretrained on the STL10 and ImageNet-1K datasets, as shown in Section~\ref{subsection:distillation}. Then, the lower-complexity distilled model is used for downstream tasks such as classification. 
\section{Results}
\label{section:results}

In this section, we demonstrate the performance of the proposed method for two downstream tasks: (a) SSL-based classification and (b) knowledge distillation-based~classification.

\subsection{SSL-Based~Classification}

We evaluate the representations obtained after pretraining the ResNet50 backbone with {\papername} on the ImageNet-1K and STL10 datasets for 1000 epochs using the loss function described above. The~model pretrained on ImageNet-1K is evaluated with a linear protocol and a semi-supervised protocol with 1\% and 10\% of images labeled.

\subsubsection{Evaluation on~ImageNet-1K}

{\bf {Linear} 
 evaluation protocol}: A linear classifier is trained on top of the frozen representations of the ResNet50~\cite{he2016deep} pretrained using {\papername} for 100 epochs with the cross-entropy~loss. 

{\bf {Semi-supervised} 
 evaluation protocol}: The pretrained ResNet50 is fine-tuned with a fraction of the ImageNet-1K dataset---1\% or 10\% of sampled labels for 100 epochs with the cross-entropy~loss.

The results on the validation set of ImageNet-1K for linear and semi-supervised evaluation protocols of the model are shown in Table~\ref{table:ImageNet_1k_evaluation}.  The main advantage of the {\papername} is that it presents a new way to regularize latent space for self-supervised pretraining by using distance correlation between the embeddings from the model and hand-crafted image features. Due to the lack of computational resources, we did not run the parameter optimization for ImageNet-1K pretraining, so we think that the results could be further~improved.

\begin{table}[H]

\centering
\caption{{\bf {Classification} accuracy. Evaluation on ImageNet-1K}. Evaluation of the representations from ResNet50 {\bf {non-contrastive} 
} backbones pretrained with {\papername} on (1) linear evaluation protocol on top of frozen representations from ImageNet; (2) semi-supervised classification on top of fine-tuned representations, with 1\% and 10\% of ImageNet samples labeled.
Top-1 refers to the accuracy of a classifier by determining if the highest-probability prediction is correct, and~Top-5 refers to whether the correct answer is among the five highest probability predictions.}
\label{table:ImageNet_1k_evaluation}

\setlength{\tabcolsep}{0.49cm}
\begin{tabular}{lcccccc}
\toprule
\multirow{3.8}{*}{\textbf{Method}} & \multicolumn{2}{c}{\textbf{Linear}} & \multicolumn{4}{c}{\textbf{Semi-Supervised}}\\ \cmidrule{2-7} 
& \multirow{2.5}{*}{\textbf{Top 1}} & \multirow{2.5}{*}{\textbf{Top 5}} & \multicolumn{2}{c}{\textbf{Top 1}} & \multicolumn{2}{c}{\textbf{Top 5}} \\ \cmidrule{4-7} 
 & & & \textbf{1\%} & \textbf{10\%} & \textbf{1\%} & \textbf{10\%} \\ \midrule
Supervised & 76.5 & - & 25.4 & 56.4 & 48.4 & 80.4 \\ \midrule
PIRL~\cite{misra2020self} & 63.6 & - & - & - & - & - \\
SimSiam~\cite{chen2021exploring} & 71.3 & - & - & - & - & - \\
InfoMin Aug~\cite{tian2020makes} & 73.0 & 91.1 & - & - & - & - \\
OBoW~\cite{gidaris2021obow} & 73.8 & - & - & - & - & - \\
BYOL~\cite{grill2020bootstrap} & 74.3 & 91.6 & 53.2 & 68.8 & 78.4 & 89.0 \\
Barlow Twins~\cite{zbontar2021barlow} & 73.2 & 91.0  & 55.0 & 69.7 & 79.2 & 89.3 \\
VICReg~\cite{bardes2021vicreg} & 73.2 & 91.1 & 54.8 & 69.5 & 79.4 & 89.5 \\
{\bf {\papername} {(ours)} 
} & \bf{{74.5} 
} & \bf{92.1} & \bf{56.1} & \bf{69.9} & \bf{79.4} & \bf{89.5} \\ \bottomrule
\end{tabular}
\end{table}
\unskip

\subsubsection*{Evaluation of Small Datasets}

In this study, we demonstrate the self-supervised learning model performance on small-scale datasets. The~model is trained on the STL10 and CIFAR20~\cite{van2020scan} datasets  with hand-crafted features: (i) flattened original images, (ii) augmented images, (iii) ScatNet features, (iv)~HOG features, and~(v) LSD features. The~proposed model achieves state-of-the-art results in the linear evaluation protocol on the STL10 and~CIFAR20 datasets compared to all other self-supervised methods. The~results for STL10 are reported in Table~\ref{table:stl10_evaluation}, and those for CIFAR20 are in Table~\ref{table:cifar20_evaluation}.

\begin{table}[H]
\centering
\caption{{\bf {Evaluation} 
 on STL10}. Classification accuracy for the linear evaluation protocol on top of frozen representations from the STL10~dataset.}
\label{table:stl10_evaluation}
\setlength{\tabcolsep}{2.58cm}
\begin{tabular}{lc}
\toprule
\textbf{Method}       &\textbf{ STL10} \\ \midrule
ADC~\cite{haeusser2018associative} & 53.0 \\
IIC~\cite{ji2019invariant} & 61 \\
TSUK~\cite{han2020mitigating} & 66.5   \\
SCAN~\cite{van2020scan} & 80.9   \\
ScatSimCLR~\cite{kinakh2021scatsimclr} & 85.1  \\
RUC~\cite{park2021improving} & 86.7   \\
{\bf {\papername} {(ours)}
} & {\bf {89.7} 
}  \\ \bottomrule
\end{tabular}
\end{table}
\unskip

\begin{table}[H]
\centering
\caption{{\bf {Evaluation} 
 on CIFAR20}. Classification accuracy for the linear evaluation protocol on top of frozen representations from the CIFAR20~dataset.}
\label{table:cifar20_evaluation}
\setlength{\tabcolsep}{2.36cm}
\begin{tabular}{lc}
\toprule
\multicolumn{1}{c}{\textbf{Method}} & \textbf{CIFAR20} \\ \midrule
IIC~\cite{ji2019invariant} & 25.7    \\
TSUC~\cite{han2020mitigating} & 35.3    \\
SCAN~\cite{van2020scan} & 50.7    \\
RUC~\cite{park2021improving} & 54.3    \\
LFER Ensemble~\cite{chong2022loss} & 56.1    \\
ScatSimCLR~\cite{kinakh2021scatsimclr} & 63.8    \\
{\bf {\papername} {(ours)} 
} & {\bf {73.2} 
}    \\ \bottomrule
\end{tabular}
\end{table}
\unskip

\subsubsection*{Transfer Learning}

To evaluate the pretrained representation of multiclass classification on the VOC07~\cite{pascalvoc2007} dataset, we train a linear classifier on top of the frozen representations from the pretrained encoder for 100 epochs. The~mAP on the VOC07 dataset is reported in Table~\ref{table:voc07_transfer}, along with results from other non-contrastive state-of-the-art SSL methods with a ResNet50 backbone.

\begin{table}[H]
\centering
\caption{{\bf {Transfer} 
 learning on multiclass classification on the VOC07~\cite{pascalvoc2007} dataset}. Evaluation of the non-contrastive representations from the pretrained model on multiclass classification using the linear classifier on top of frozen representations. We report~mAP.}
\label{table:voc07_transfer}
\setlength{\tabcolsep}{1.99cm}
\begin{tabular}{lc}
\toprule
\multirow{2}{*}{\textbf{Method}} & \textbf{Linear Classification} \\ \cmidrule{2-2} 
 & \textbf{VOC07} \\ \midrule
Supervised & 87.5 \\ \midrule
PIRL~\cite{misra2020self} & 81.1 \\
BYOL~\cite{grill2020bootstrap} & 86.6 \\
OBoW~\cite{gidaris2021obow} & \bf{{89.3} 
} \\
Barlow Twins~\cite{zbontar2021barlow} & 86.2 \\
VICReg~\cite{bardes2021vicreg} & 86.6 \\
{\bf \papername {(ours)} 
} & 87.1 \\ \bottomrule
\end{tabular}
\end{table}
\unskip

\subsection{Knowledge Distillation-Based~Classification}
\label{subsection:distillation}

To evaluate the proposed approach on the knowledge distillation-based classification task, we have used a pretrained CLIP~\cite{radford2021learning} model based on the ViT-B-16~\cite{dosovitskiy2020vit} encoder as the teacher and ResNet50~\cite{he2016deep} as the student model. The~CLIP model is trained based on the contrastive loss between the image and text embeddings. To~proceed with the knowledge distillation in the same way as the SSL training, we use the default projector 8192-8192-8192 after the ResNet50 encoder. The~pretrained CLIP ViT model uses images of shape $224 \times 224 \times 3$ as an input and outputs a latent vector of shape 512, as~shown in Figure~\ref{fig:mv_mr_distill}. When reporting the results, the~teacher model is evaluated using zero-shot evaluation on the ImageNet-1k dataset and a linear evaluation pipeline on other datasets. The~student model is evaluated using a linear evaluation pipeline on all~datasets.

The goal of the experimental validation is to demonstrate whether the ResNet50 model with 23.5 M  parameters trained only on a smaller dataset can provide similar performance to the CLIP based on the ViT model with 86.2 M parameters and trained on 400 M images. It is important to point out that the training is performed without any additional labels, according to the proposed knowledge distillation framework. In~Table~\ref{table:distillation_imagenet1k}, we report results of knowledge distillation, where CLIP based on ViT-B-16 is used as a teacher model and ResNet50 is used as a student model. The~model is trained for only 200~epochs on a single NVIDIA RTX2080ti GPU using the proposed knowledge distillation approach on the STL10, CIFAR100, and~ImageNet-1K datasets. The~obtained results confirm that the convolutional ResNet50 model with 4$\times$ fewer parameters in comparison to the transformer ViT teacher model and trained on a considerably smaller amount of unlabeled data can closely approach the performance of the teacher model without any special labeling, clustering, additional augmentations, or~complex contrastive losses. Remarkably, the~proposed knowledge distillation largely preserved this performance and achieved 95.6\%  versus the best SSL {\name} result of 89.7\%,  as indicated in Tables~\ref{table:stl10_evaluation} and \ref{table:distillation_imagenet1k}.  The~proposed distillation method outperforms all other distillation methods on the CIFAR100 dataset: {\bf {78.6\%} 
} vs. current state-of-the-art 78.08\% \cite{chen2022knowledge}. Thus, both the proposed {\name} SSL training and knowledge distillation achieve state-of-the-art results on the STL10 and CIFAR100 datasets and demonstrate competitive results for the ImageNet-1K among all non-contrastive and clustering-free SSL~methods.

\begin{table}[H]
\centering
\caption{Knowledge distillation experiment with CLIP based on ViT-B-16 as teacher model and~ResNet50 as a student~model.}
\label{table:distillation_imagenet1k}
\setlength{\tabcolsep}{0.42cm}
\begin{tabular}{lcccc}
\toprule
\textbf{Approach} &  \textbf{Parameters}  & \textbf{STL10} & \textbf{ImageNet-1K} & \textbf{CIFAR100} \\ \midrule
CLIP ViT-B-16 \\ (zero-shot) &  86.2 M   & -   & 67.1  & -                 \\  \midrule
CLIP ViT-B-16 \\ (linear evaluation) & 86.2 M & 98.5  & 77.4 &  82.2                    \\  \midrule
{\bf {\name} 
} ResNet50 \\ (linear evaluation) & 23.5 M & 95.6 & 75.3 &  78.6                 \\ \bottomrule
\end{tabular}
\end{table}
\unskip

\subsection{Ablation~Studies}
\label{section:ablation_studies}

In this subsection, we describe the ablation studies on the proposed losses (Table~\ref{table:loss_ablations}).  In~each of the experiments, we use the same training and evaluation setup: dataset---STL10, epochs---$100$, batch size---$64$, 16-bit precision, batch accumulation---$1$ batch. We use a linear evaluation pipeline. We demonstrate the impact of representation learning based on the maximization of the considered upper bound on the mutual information and the maximization of distance covariance in various settings. In~this ablation, we show that the best results are achieved when using three loss terms: $\mathcal{L}_1$, $\mathcal{L}_2$, and~$\mathcal{L}_3$. 

\begin{table}[H]
\centering
\caption{{\bf {Ablation studies on the combination of losses}
}. We check the importance of each loss term for the training of the model. It is shown that using loss terms $\mathcal{L}_{1}$, $\mathcal{L}_{2}$, and $\mathcal{L}_{3}$ provides the best classification performance. Also, we observe a phenomenon in which loss terms $\mathcal{L}_{2}$ and $\mathcal{L}_{3}$ work the best when applied jointly with the loss $\mathcal{L}_{1}$. However, it is interesting to point out that a disjoint usage of these losses does not lead to reasonable performance enhancement. The~exact nature of this phenomenon is not completely clear and additional investigation should be~performed.}
\label{table:loss_ablations}
\setlength{\tabcolsep}{1.06cm}
\begin{tabular}{ccccc}
\toprule & & & \multicolumn{2}{c}{\textbf{Accuracy}} \\ \cmidrule{4-5} 
\multirow{-2}{*}{ \boldmath{$\mathcal{L}_1$}} & \multirow{-2}{*}{\boldmath{$\mathcal{L}_2$}} & \multirow{-2}{*}{\boldmath{$\mathcal{L}_3$}} & \textbf{Top 1} & \textbf{Top 5}                        \\ \midrule
\multicolumn{5}{c}{1 loss}                                                                                \\ \midrule
\checkmark & & & 50.86 & 93.95  \\
 & \checkmark & & 46.71 & 92.18 \\
 & & \checkmark & 44.1  & 92.08 \\ \midrule
\multicolumn{5}{c}{2 losses} \\ \midrule
\checkmark & \checkmark & & 50.76 & 93.83 \\
\checkmark & & \checkmark & 47.39 & 92.54 \\
 & \checkmark & \checkmark & 40.06 & 89.31 \\ \midrule
\multicolumn{5}{c}{3 losses} \\ \midrule
\checkmark & \checkmark & \checkmark & {\bf {69.38}
} & {\bf {98.85} 
} \\ \bottomrule
\end{tabular}
\end{table}
\unskip

\section{Implementation~Details}
\label{appendix:implementation}

The architecture of the {\papername} is similar to ones used in other SSL methods such as BarlowTwins~\cite{zbontar2021barlow}, VICReg~\cite{bardes2021vicreg}, and~others. The~model $f_{\phi_{z}}$, shown in Figure~\ref{fig:d_cor_ssl}, consists of two main parts: (i) the encoder, which is used for downstream tasks, and~(ii) the projector, which is used for the mapping of encoder outputs to the embeddings used for the training loss functions in (\ref{fig:d_cor_ssl}). In~our experiments, we use standard ResNet50~\cite{he2016deep}, available in the {$torchvision$} 
 library~\cite{paszke2019pytorch}, as the encoder and projector, which consists of two linear layers of size $8192$, followed by batch normalization, ReLU, and output linear~layer.

We use computer vision feature-extraction methods applied to the original data: original RGB image (that is being flattened into a feature vector), ScatNet features of the image~\cite{andreux2020kymatio}, randomly augmented images, flattening into a feature vector, histogram of oriented gradients (HOG), and~local standard deviation filter (LSD filter) \cite{narendra1981real}.

\textbf{{ScatNet}
 transform}: ScatNet~\cite{oyallon2018scattering, andreux2020kymatio} is a class of Convolutional Neural Networks (CNNs) that have a set of useful properties: (i) deformation stability, (ii) fixed weights, (iii)~sparse representations, (iv) interpretable~representation.

\textbf{{Randomly} 
 augmented image}: In our experiments, we have applied the following augmentations to the image: random cropping, horizontal flipping, random color augmentations, grayscale, and Gaussian blur. Then, the image is flattened into a one-dimensional feature~vector.

\textbf{{HOG}
}: Histogram of oriented gradients (HOG) \cite{dalal2005histograms} is a feature description that is based on the counting of occurrences of gradient orientation in the localized portion of an~image. 

\textbf{{LSD} 
 filter}: A local standard deviation filter~\cite{narendra1981real} is a filter that computes a standard deviation in a defined image region over the image. The~region is usually of a rectangular shape of size $3 \times 3$ or $5 \times 5$ pixels.

We use the PyTorch framework~\cite{paszke2019pytorch} for the implementation of the proposed approach.
We use ScatNet with the following parameters: $J = 2$ and $L = 8$. We use the HOG feature extractor with the following parameters: number of bins---$24$ and pool size ---$8$. We use a kernel of size $3\times 3$ in the STD filter. As~augmentations, for both image representation and as the input to the encoder, we use randomly resized cropping; random horizontal flipping with probability $0.5$; random color-jittering augmentation with brightness $0.8$, contrast $0.8$, saturation $0.8$, hue $0.2$, and probability $0.8$; random grayscale with probability $0.2$; and Gaussian blur with a kernel size of $0.1$ of the image size, mean $0$, and sigma in the range $[0.1, 2]$.

For the losses, the~margin parameter $\gamma$ is set to $1$, and~$\epsilon$ is set to $1e-4$ in (\ref{equation:Variance_reg}).

During the {\bf {self-supervised} 
 pretraining} experiments that are presented in Table~\ref{table:ImageNet_1k_evaluation} and Table~\ref{table:stl10_evaluation}, we train models for $1000$ epochs, with batch size $256$, gradient accumulation every 4 steps, base learning rate $1e-4$, Adam~\cite{kingma2014adam} optimizer, cosine learning rate schedule, and~$16$-bit precision. During~{\bf {linear} 
 evaluation}, we train a single-layer linear model for $100$ epochs with batch size $256$, learning rate $1e-4$, and~Adam optimizer. During~{\bf {semi-supervised} 
 evaluation} on ImageNet-1K, we train a model for $100$ epochs with batch size $128$, learning rate $1e-4$, and~Adam optimizer. During~the knowledge distillation, we train the model for $200$ epochs, with batch size~$512$, base learning rate $1e-4$, Adam optimizer, cosine learning rate schedule, and~$16$-bit~precision.

When training, weight parameters $\lambda = 1$ and  $\mu = 1$ in in $\mathcal{L}_{1}$, $\alpha = 1$ in $\mathcal{L}_{2}$, and $\beta_{k} = 1, k = 1...K$ in  $\mathcal{L}_{3}$.

\section{Conclusions}

In this paper, we introduce novel self-supervised {\papername} learning and knowledge distillation approaches, which are based on the maximization of several dependency measures between two embeddings obtained from views with and without augmentations and multiple representations extracted from non-augmented views. The~proposed methods use an upper bound on mutual information and a distance correlation for the dependence estimation for the representations of different dimensions. We explain the intuition behind the proposed method of upper bound on the mutual information and the usage of distance correlation as a dependence measure. Our method achieves state-of-the-art self-supervised classification on the STL10 and CIFAR20 datasets and comparable state-of-the-art results on ImageNet-1K datasets in linear evaluation and semi-supervised evaluations. We show that ResNet50 pretrained using knowledge distillation on CLIP ViT-B-16 achieves comparable performance with far fewer parameters (23.5 M with ResNet50 vs. 86.2 M parameters with CLIP ViT-B-16) and a relatively small training set on multiple datasets: STL10 and ImageNet-1k. The~proposed disillation method also achieves state-of-the-art peformance on the CIFAR100 dataset, 78.6\% vs. previous state-of-the-art of 78.08\%.

In our paper, we exclusively focus on the image-classification downstream task using ResNet architecture, limiting ourselves to its standard augmentation techniques. Future efforts should extend beyond ResNets to encompass transformers and other advanced deep learning architectures, exploring their applicability not just in classification but also in other vision downstream tasks, such as object-detection and -segmentation tasks, by~using the pretrained backbone with a proper head that is finetuned for the selected downstream task. This expansion would allow for a broader range of augmentation strategies, such as Masked Image Modeling (MIM), and~provide insights into the performance of different hand-crafted features across various architectures, enhancing the versatility of self-supervised learning approaches in computer~vision.
The code is available at: \href{https://github.com/vkinakh/mv-mr}{github.com/vkinakh/mv-mr}.

\appendix
\section{Ablation Studies}

In this section, we describe the ablation studies on the combination of features for loss term $\mathcal{L}_3$ (Table \ref{table:features_ablations_zz}), a~number of layers, and both the~size of the projector in the trainable encoder (Table \ref{table:projector_ablations}) and~image augmentations (Table \ref{table:augmentation_ablations}). In~each of the experiments, we use the same training and evaluation setup: dataset---STL10~\cite{coates2011analysis}, epochs---$100$, batch size---$64$, 16-bit precision, batch accumulation---$1$ batch. When pretraining, all three loss terms are used. After~model pretraining, it is evaluated using linear~evaluation.

We describe the ablation studies on the combinations of features used for the $\mathcal{L}_3$ loss term in combination with loss terms $\mathcal{L}_{1}$ and $\mathcal{L}_{2}$ in Table~\ref{table:features_ablations_zz}. We study the impact of features on the classification accuracy of the model. We use the following features in the study: the original image flattened into a vector, ScatNet~\cite{andreux2020kymatio} features of the original image, an~augmented image flattened into a vector, a histogram of oriented gradients of the original image, and~features from the local standard deviation filter (LSD). We use ScatNet with the following parameters: $J = 2$ and $L = 8$. We use the HOG~\cite{dalal2005histograms} feature extractor with the following parameters: number of bins---$24$ and pool size---$8$. A kernel of size $3\times3$ is used in the LSD filter~\cite{narendra1981real}. As~augmentations for image representation, we use randomly resized cropping; random horizontal flipping with probability $0.5$; random color-jittering augmentation with brightness $0.8$, contrast $0.8$, saturation $0.8$, hue $0.2$, and probability $0.8$; random grayscale with probability $0.2$; and Gaussian blur with a kernel size of $0.1$ of the image size, mean $0$, and sigma in the range $[0.1, 2]$. We show that the best results are achieved when we use the combination of all feature extractors mentioned~above.

The ablation studies on image augmentations are presented in Table~\ref{table:augmentation_ablations}. As~augmentations, we compare 
randomly resized cropping, random horizontal flipping,  random color augmentations, random grayscale, and~random Gaussian blur. We use the same parameters for each augmentation as~when augmented images are used as features. We show that the best classification results are achieved when a combination of random cropping, horizontal flipping, color jittering, and~random grayscale is~used.

The ablation studies on the number of layers and their size in the encoder's projects are presented in Table~\ref{table:projector_ablations}. When the number of layers is larger than one in the projector, it consists of blocks with linear layers, batch normalization, and~ReLU activation. We always keep the last layer linear. We show that the best classification results are observed when the projector consists of three layers, each with 8192-8192-8192 neurons.

\begin{table}[H]
\small
\centering
\caption{Ablation studies of the combinations of features used for the $\mathcal{L}_3$ loss. In~this setup, all three losses are used: $\mathcal{L}_1$, $\mathcal{L}_2$, and $\mathcal{L}_3$. ScatNet transformation of the original image. Augmented image---randomly augmented original image. HOG---histogram of oriented gradients, computed from original view. LSD---original view filtered with the local standard deviation filter. Since all of these features are images, they are flattened before computing distance~correlation.}
\label{table:features_ablations_zz}
\setlength{\tabcolsep}{3.45mm}
\begin{tabular}{ccccccc}
\toprule
\multirow{2}{*}{\textbf{Original Image}} & \multirow{2}{*}{\textbf{ScatNet}} & \multirow{2}{*}{\textbf{Augmented Image}} & \multirow{2}{*}{\textbf{HOG}} & \multirow{2}{*}{\textbf{LSD}} & \multicolumn{2}{c}{\textbf{Accuracy}} \\ \cmidrule{6-7} 
& & & & & \textbf{Top 1} & \textbf{Top 5}        \\ \midrule
\multicolumn{7}{c}{1 feature}  \\ \midrule
\checkmark & & & & & 58.82         & 96.81        \\
 & \checkmark & & & & 54.12        & 95.23        \\
 & & \checkmark & & & 63.51        & 97.81        \\
 & & & \checkmark & & 54.15        & 95.26        \\
 & & & & \checkmark & 53.94        & 95.44        \\ \midrule
\multicolumn{7}{c}{2 features}                    \\ \midrule
\checkmark & \checkmark & & & & 64.44 & 97.97     \\
\checkmark & & \checkmark & & & 66.18 & 98.39     \\
\checkmark & & & \checkmark & & 63 & 97.78        \\
\checkmark & & & & \checkmark & 63.14 & 97.8      \\
 & \checkmark & \checkmark & & & 63.3 & 97.78     \\
 & \checkmark & & \checkmark & & 62.95 & 97.6     \\
 & \checkmark & & & \checkmark & 59.41 & 96.78    \\
 & & \checkmark & \checkmark & & 63.66 & 97.69    \\
 & & \checkmark & & \checkmark & 60.21 & 96.8     \\
 & & & \checkmark & \checkmark & 62.46 & 97.71    \\ \midrule
\multicolumn{7}{c}{3 features} \\ \midrule
\checkmark & \checkmark & \checkmark & & & 65.82 & 98.18   \\
\checkmark & \checkmark & & \checkmark & & 65.52 & 97.97   \\
\checkmark & \checkmark & & & \checkmark & 60.96 & 97.08   \\
\checkmark & & \checkmark & \checkmark & & 65.11 & 98.12   \\
\checkmark & & \checkmark & & \checkmark & 65.19 & 98      \\
\checkmark & & & \checkmark & \checkmark & 65.37 & 98.29   \\
 & \checkmark & \checkmark & \checkmark & & 65.45 & 98.18  \\
 & \checkmark & \checkmark & & \checkmark & 64.35 & 97.93  \\
 & \checkmark & & \checkmark & \checkmark & 60.63 & 97.1   \\
 & & \checkmark & \checkmark & \checkmark & 64.9  & 98.08  \\ \midrule
\multicolumn{7}{c}{4 features} \\ \midrule
\checkmark & \checkmark & \checkmark & \checkmark & & 68.25 & 98.45 \\
\checkmark & \checkmark & \checkmark & & \checkmark & 68.2  & 98.53 \\
\checkmark & \checkmark & & \checkmark & \checkmark & 64.56 & 97.44 \\
\checkmark & & \checkmark & \checkmark & \checkmark & 67.21 & 98.48 \\
 & \checkmark & \checkmark & \checkmark & \checkmark & 67.05 & 98.22 \\ \midrule
\multicolumn{7}{c}{5 features} \\ \midrule
\checkmark & \checkmark & \checkmark & \checkmark & \checkmark & {\bf {69.38} 
} & {\bf {98.85} 
} \\ \bottomrule
\end{tabular}

\end{table}
\unskip

\begin{table}[H]
\centering
\caption{Ablation studies on the image~augmentations.}
\label{table:augmentation_ablations}
\setlength{\tabcolsep}{3mm}
\begin{tabular}{ccccccc}
\toprule
\multirow{2}{*}{\textbf{Random Crop}} & \multirow{2}{*}{\textbf{Horizontal Flip}} & \multirow{2}{*}{\textbf{Color}} & \multirow{2}{*}{\textbf{Grayscale}} & \multirow{2}{*}{\textbf{Blur}} & \multicolumn{2}{c}{\textbf{Accuracy}} \\ \cmidrule{6-7} & & & & & \textbf{Top 1} & \textbf{Top 5} \\ \midrule
\checkmark & & & & & 43.02 & 90.31 \\
\checkmark & \checkmark & & & & 49.99 & 93.98 \\
\checkmark & \checkmark & \checkmark & & & 50.58 & 93.76 \\
\checkmark & \checkmark & \checkmark & \checkmark & & {\bf {70.82} 
} & {\bf {98.96}
} \\
\checkmark & \checkmark & \checkmark & \checkmark & \checkmark & 69.38 & 98.85 \\ \bottomrule
\end{tabular}

\end{table}

\begin{table}[H]
\centering
\caption{Ablation studies on the projector. Projectors consist of blocks with linear layers, batch normalization, and~ReLU activation. We always keep the last layer~linear.}
\label{table:projector_ablations}
\setlength{\tabcolsep}{24.1mm}
\begin{tabular}{lr}
\toprule
\multicolumn{1}{c}{\textbf{Projector Size}} & \multicolumn{1}{c}{\textbf{Accuracy}} \\ \midrule
\textbf{{8192-8192-8192}
}                     & \textbf{{69.38} 
}                        \\
4096-4096-4096                              & 51.90                                  \\
2048-2048-2048                              & 51.35                                 \\
1024-1024-1021                              & 51.86                                 \\
512-512-512                                 & 49.40                                 \\
256-256-256                                 & 49.02                                 \\
8192-8192                                   & 49.90                                  \\
4096-4096                                   & 48.81                                 \\
2048-2048                                   & 48.66                                 \\
1024-1024                                   & 48.73                                 \\
512-512                                     & 48.06                                 \\
256-256                                     & 48.07                                 \\
8192                                        & 48.65                                 \\
4096                                        & 48.51                                 \\
2048                                        & 48.20                                  \\
1024                                        & 47.61                                 \\
512                                         & 46.11                                 \\
256                                         & 47.17                                 \\
without projector                           & 16.70                                  \\ \bottomrule
\end{tabular}

\end{table}


\normalsize
\bibliography{references}


\end{document}